\documentclass[letterpaper, 10 pt, conference]{ieeeconf}  

\makeatletter
\def\endthebibliography{%
	\def\@noitemerr{\@latex@warning{Empty `thebibliography' environment}}%
	\endlist
}
\makeatother

\IEEEoverridecommandlockouts                              

\usepackage{graphicx} 
\usepackage{eurosym}
\usepackage{cite}
\usepackage[table]{xcolor}

\title{\LARGE \bf
	Including Sparse Production Knowledge into Variational Autoencoders to Increase Anomaly Detection Reliability
}

\author{Tom Hammerbacher$^{1,2}$, Markus Lange-Hegermann$^{1}$, Gorden Platz$^{2}$
	\thanks{$^{1}$Department of Electrical Engineering and Computer Science, Ostwestfalen-Lippe University of Applied Sciences, Campusallee 12, 32657 Lemgo, Germany
		{\tt\small markus.lange-hegermann@th-owl.de}}%
	\thanks{$^{2}$Phoenix Contact Electronics GmbH, Dringenauer Str. 30, 31812 Bad Pyrmont, Germany
		{\tt\small thammerbacher@phoenixcontact.com, gplatz@phoenixcontact.com}}%
}

\begin{document}
			
	\maketitle
	\thispagestyle{empty}
	\pagestyle{empty}
		
	\begin{abstract}
	
	Digitalization leads to data transparency for production systems that we can benefit from with data-driven analysis methods like neural networks. For example, automated anomaly detection enables saving resources and optimizing the production. We study using rarely occurring information about labeled anomalies into Variational Autoencoder neural network structures to overcome information deficits of supervised and unsupervised approaches. This method outperforms all other models in terms of accuracy, precision, and recall. We evaluate the following methods: Principal Component Analysis, Isolation Forest, Classifying Neural Networks, and Variational Autoencoders on seven time series datasets to find the best performing detection methods. We extend this idea to include more infrequently occurring meta information about production processes. This use of sparse labels, both of anomalies or production data, allows to harness any additional information available for increasing anomaly detection performance.
		
	\end{abstract}

	\section{INTRODUCTION}
	
	In the last years, digitalization and the Internet of Things created a vast information basis, also called Big Data. This promotes new solutions for the economy and society, such as remote home and work models, self-driving transportation, cloud storage, online payment, and social networks \cite{Daily.2017, Antonopoulos.2010}. In the fourth industrial revolution, new production processes are designed to become self-optimizing and autonomous. Automated anomaly detection is one example of digitalizing production facilities to increase productivity \cite{HenningKagermannJohannesHelbigArianeHellingerWolfgangWahlster.2013,A.Alvanpour.2020}. Machine learning approaches show satisfying results on system failure detection for solutions with available information about machine failures (supervised) or without labeled anomaly information (unsupervised) \cite{Omar.2013,Z.Wang.2020}. In one example, such solutions save 870,000 EUR for every shortened downtime on average by increasing the availability \cite{VansonBourne.2020}. Furthermore, data-driven detection performs well without additional sensors or human interaction.
	
	Therefore, companies gather process data, which is easy to acquire, because sensor driven, automated systems monitor and compute this information already. In contrast, labeling anomalous production facility behavior or defective products is difficult because of the resource-intensive process and rare occurrence. For example, to track production process reliably, one needs to check all products after each production step to connect the anomaly label with the corresponding data chunk. Domain experts must  label the anomalies manually without access to an automatic fault detection system. That is why there are only few anomaly labels in comparison to the process data. One says that these labels are sparse in the data.
	
	On this basis, most anomaly detection algorithms do not rely on anomaly labels, i.e., they employ deviation based, unsupervised detection methods and leave the anomaly labels for validation only \cite{Wang.2018}. If an algorithm learns to model the anomaly free production process, a deviation from ideal condition leads to a difference in the modeling process that gets detected. An alternative is to create a supervised model based on process data for which experts label anomalies \cite{Goernitz.2013}. Similar anomalies result in similar process data, which we can classify with our model. 
	
	Both procedures disregard information that could be highly beneficial for anomaly detection. Unsupervised methods cannot learn from the labeled anomalies whereas supervised methods cannot learn from unlabeled process data.
	
	To combine the information in labeled and unlabeled data, we construct a model that can handle both unlabeled time series and connect it with specific sparse anomaly labels. In practice, we combine an unsupervised Variational Autoencoder (VAE) with supervised neural network techniques inspired by \cite{Berkhahn.2019}. Our model reconstructs the process data and predicts the sparse labels inside the model's bottleneck. Moreover, we reconstruct the input from the latent representation and the predicted labels to bring domain knowledge into the decoder. Our inclusion of sparse anomaly knowledge leads to better performing models. Our extended VAEs can be trained on more data and make use of combining different predictions for the detection target like neural network ensembles do \cite{Hansen.1990}. Besides, optimizing towards one training target leads to an improvement of the other targets, called transfer learning. As a bonus, we can use the sparse anomaly label predictions to enhance the anomaly detection in parallel.
	
	We compare our model to other state-of-the-art anomaly detection methods like Variational Autoencoders \cite{Doersch.2016}, Classifying Neural Networks \cite{Goh.2017}, Isolation Forest \cite{Liu.2008}, and Principal Component Analysis \cite{Wold.1987} in terms of performance metrics like f1-score and AUC-score as well as manageability. We additionally use the reconstruction probability \cite{An.2015} instead of the reconstruction error to include probabilistic domain knowledge about the process data. This allows to distinguish process noise from anomalies.
	
	In this paper, section \ref{DataBasis} introduces the observed datasets and the anomaly detection procedure. Our machine learning model and training are explained in section \ref{MODEL_TRAINING}. Section \ref{RESULTS} compares our results. We conclude in section \ref{CONCLUSION}.

	\section{DATA BASIS AND EXPERIMENTAL SETUP}
	\label{DataBasis}
	
	To compare the performances of our models, we need a clear understanding of anomalies. We extend the definition from \cite{Hawkins.1980} to an observation with a negative impact on the production process, which deviates from routine behavior because another process is responsible for the outcome. We monitor the optimization on the training dataset with a validation dataset to spot problems like overfitting. Both datasets are free of anomalies to ensure that the unsupervised methods learn good representations of the anomaly free process. For the supervised training, we need a dataset with anomalous and anomaly free data. Afterwards, we compare the anomaly classification performances of all trained models on a testing dataset with and without anomalies. To create these mixed datasets, we separate time series with a defined amount of anomalies by taking all even rows for testing and all odd rows for supervised training. This is typical for online learning, and might bias the results in a positive way. Next, we duplicate every row to ensure the correct time dependencies compared to the unsupervised training and validation datasets. This corresponds to decreasing the resolution of a picture while keeping the dimensions, just as time series. We repeat every experiment with the original time series and spot performance differences that are biased because of the data set manipulation. We found no significant deviation between both testing sets. Final dataset sizes and the proportion of anomalies can be observed in Table \ref{Table:Dimensions}. The defined amount of anomalies for every dataset is given in brackets.
		
	\begin{table}[h]
	\caption{Dataset sizes for training unsupervised (US Training) and supervised (S Training), validation, and Testing. The percentage of anomalies besides zero is given in brackets}
	\label{Table:Dimensions}
	\begin{center}
	\begin{tabular*}{\linewidth}{@{\extracolsep{\fill}}|c|c|c|c|c|}	
		\hline
		Dataset & Features & US Training & Validation & S Train \& Test\\
		\hline 
		Con&18&7,900&800&2,400 [12\%]\\
		\hline
		DB&6&1,700&100&2,600 [63\%]\\
		\hline
		Gen&18&9,300&1,000&3,200 [15\%]\\
		\hline
		Pneu&7&272,000&254,000&375,000 [32\%]\\
		\hline
		Sol&22&6,201,000&688,000&237,000 [7\%]\\
		\hline
		Stor&18&17,200&1,900&19,600 [23\%]\\
		\hline
		UPS&18&10,700&1,100&670 [68\%]\\
		\hline
	\end{tabular*}
	\end{center}
	\end{table}

	We consider 7 datasets. The internal "Sol" dataset contains 22 continuous, electrical features from 7,120,000 power consumption records of a selective soldering machine from Phoenix Contact GmbH \& Co. KG (Phoenix Contact). We preprocess four production shifts with a production status label (production, equip, rest) as metadata out of a total of 66 shifts. We create artificial anomalies based on domain experts' suggestions with random noise on specific features for evaluation. The internal "UPS" dataset from Phoenix Contact consists of 12,000 anomaly free training records and 700 testing data points with 68 percent of artificial and natural created anomalies \cite{Muller.2020}. The 18 continuous variables are generated by an uninterruptable power supply with a consumption system. For the internal "Pneu" dataset from Emerson Automation Solutions - AVENTICS GmbH, we monitor the airflow and six binary valve switching signals of a pneumatic system in the frequency domain. We investigate 530,000 training data points without anomalies as well as 8,000,000 data points from 12 different leakage system setups. The "DB" dataset contains soft motion axis data artificially created by a programmable logic controller. Different axis timings influence six features for 4,400 records \cite{Niggemann.2018}. For the "Stor" dataset, we evaluate 38,000 data points with 18 features from a high rack storage system \cite{Niggemann.2018}. We also analyze the "Gen" SmartFactoryOWL dataset that contains 18 features and 13,000 records from a pneumatic pick-and-place demonstrator. Gripping and storage units move wood and metal pieces into predefined positions \cite{Schuster.2018}. The SmartFactoryOWL additionally provides the "Con" dataset, consisting of 11,100 records and 18 features from the power consumption and control program. Five drivers and conveyor belts allow vertical and horizontal transportation \cite{Iosbina.2020}. 
	
	In these datasets, anomalies are labeled based on expert knowledge. We scale every feature within a range between 0 and 1. Additionally, we remove features with zero variance, keeping in mind that separately monitoring them in practice can still be useful. If a model can interpret time behavior, we optimize the window size for time quantization as a hyperparameter.
	
	In terms of anomaly evaluation, we compare the following methods: Isolation Forest (IsoF), Classifying Neural Networks (DNN), Principal Component Analysis (PCA), and Variational Autoencoders with reconstruction error \cite{Doersch.2016} (VAE Err), reconstruction probability \cite{An.2015} (VAE Prob), and sparse label extensions \cite{Berkhahn.2019} with prediction averaging (VAE SL Avg) and maximum selection (VAE SL Max). For the "Sol" dataset, we create a VAE extension (VAE MD) that can handle the sparse production status information. The Isolation Forest and Classifying Neural Network implementations output the anomaly labels directly. Except for these first two methods, the anomaly detection relies on the dimensionality reduction and signal reconstruction ability. 
	
	For anomaly classification based on these signal reconstruction methods, we calculate the deviation between the input data and the models reconstruction. In detail, we calculate the mean over all features both for the input data and the reconstruction with our loss function of the model. Next, we define a threshold to classify every data point with a loss greater than this value as anomaly. We calculate this threshold with the help of the loss over the validation dataset. It is selected such that a fixed percentage of values is lower than the validation loss. We choose this percentage such that we receive the highest f1-score for our test dataset. In practice, we can adjust our decision point based on the importance of sensitivity versus specificity of the anomaly detection.
	
	We separate the true and false negatives and positives for every percentile of all deviation-based methods from 0 to 100 as well as once for the anomaly labels and classification-based methods. Next, we derive accuracy, precision, recall, f1-score, and AUC-score on our test sets. The highest f1-score cut after two decimal digits selects the best result. Equal scores lead to choosing the highest AUC-score between them. 
	
	Every final setup for each experiment, except for the Principal Component Analysis, is repeated 50 times, selecting the best result. To find the optimal hyperparameter values, we initialize all models with the constraints we explain in section \ref{MODEL_TRAINING}. We optimize our search space by analyzing over- and underfitting as well as our metrics in order to decide the next optimization step. Window sizes for time quantization are inspired by the work cycles of the analyzed processes. During training, we optimize our hyperparameters to decrease our loss functions.

	\section{MODEL AND TRAINING}
	\label{MODEL_TRAINING}
	We implement anomaly detection in time series with different supervised and unsupervised approaches. Therefore, we explain our model structure as well as hyperparameters found by extensive experiments.
	
	For comparability, we train all neural network-based techniques with the Adam optimizer using an initial learning rate of 1e-3, Glorot normal weight and zero bias initialization, and L2 regularization of 1e-3. To ensure learning stability, we decrease the learning rate down to 1e-4 if necessary. To model time dependencies, we include LSTM cells activated via tanh and recurrent sigmoid. Time distributed dense layers ensure the needed time quantization through the whole neural network. They are equipped with ReLu activation functions. Output and bottleneck layers for continuous signals are not equipped with an activation. We use sigmoid activation functions at the output and bottleneck to model probabilities of binary features. Alternatives were tested and did not change the results qualitatively.
	
	Our different VAEs enable anomaly detection based on different reconstruction abilities for anomaly free and anomalous data. We optimize them to reconstruct anomaly free data points during training. As a consequence, they fail to reconstruct anomalous data adequately which leads to a higher reconstruction loss. For our experiments, the VAE variants are built symmetrically, except for the bottleneck and probabilistic outputs described in the following section. Layers closer to the bottleneck are smaller or equal in size with the bottleneck forming the smallest width for the mean and variance layer. The number of neurons in the input and output layer is equal to the number of dataset features. The first and last hidden layer is double the input layer's width to encode more complex data context followed by up to three hidden time distributed dense layers with hyperparameters for amount and width. Around the bottleneck, we place up to three LSTM layers. We create hyperparameters for bottleneck width, batch size, and epochs to optimize them in the same way. Moreover, we optimize recurrent and time distributed layer amounts and width during the experiments. Our \textit{VAE Err} model combines the L2-norm as loss function with the KL-loss inside of the bottleneck. Both summands are weighted with a hyperparameter to ensure a valid latent space without posterior collapse and right reconstruction. 
		
	We extend the decoder structure for the \textit{VAE Prob} model by replacing the output layer with three layers fully connected with the last hidden layer. The first two output layers model a Gaussian probability distribution with mean and log variance for continuous signals, the third represents the mean of a Bernoulli distribution for binary signals. To train a valid probability distribution, we replace the reconstruction error with the calculation of a reconstruction probability, i.e., we calculate the likelihood for the input data belonging to the probability distribution of the autoencoder's prediction. In our implementation, we compute the negative log probability that the input was drawn from a normal Gaussian distribution defined by the output's mean and log variance for continuous signals and the probability that the input was drawn from a Bernoulli distribution defined by the outputs mean for discrete signals. We add hyperparameter weights to both loss functions to influence their importance. 
	
	\begin{figure}[thpb]
		\centering
		\framebox{\parbox{3in}{\includegraphics[scale=1.0]{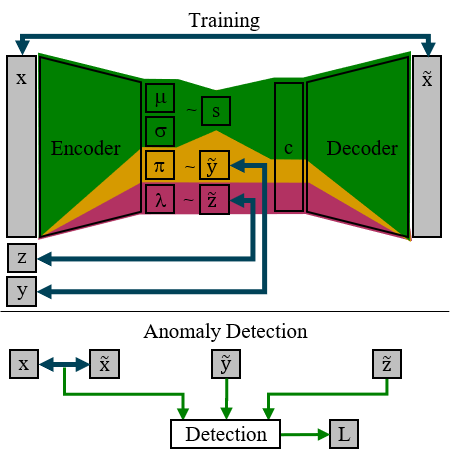}}}		
		\caption{Variational Autoencoder (VAE Prob) in green with extensions for sparse anomaly labels (VAE SL) in yellow and sparse production knowledge (VAE MD) in red. Blue arrows symbolize calculating the reconstruction deviation.}
		\label{figure:model}
	\end{figure}

	
	For our \textit{VAE SL} model, pictured in Fig. \ref{figure:model}, we use the \textit{VAE Prob} model as a basis, represented by the green figure around mean layer $\mu$, variance layer $\sigma$, and sampled representation $s$. We create an extension, colored in yellow, by adding a layer activated by a softmax function $\pi$ in parallel to the autoencoder's latent space $s$ in the bottleneck, which we concatenate with the sampled latent space for the decoder's input $c$. This additional output computes the prediction for the anomaly label $\tilde{y}$. For training, we calculate the Bernoulli likelihood between the real anomaly label and the prediction, if available. We add this term to the loss functions, again with a hyperparameter weight to trade between label prediction and signal reconstruction. 
	
	In general, our anomaly label is represented by a one dimensional time series with zeros for data without anomaly and with ones for anomalous data labeling. On this basis, we cannot model uncertainty about the occurrence of anomalies for our unlabeled data. One-hot-encoding for our anomaly label helps us to include this uncertainty. Column one represents the occurrence of anomalous data while column two represents the occurrence about data without anomalies. We set the training loss of this summand equal to zero if no anomaly label information is available. Moreover, if the current batch is labeled as anomalous for training, we ignore the reconstruction, because we only want to learn anomaly free data reconstructions. We create one more hyperparameter on this basis to individually weight the Bernoulli loss for an anomaly in comparison to the anomaly free process data for our anomaly labeling. These hyperparameters are essential to counteract the imbalance of the dataset, if anomalies are infrequent. Otherwise, the model labels all data as anomaly free to create a small loss. 
	
	Our model combines supervised and unsupervised anomaly detection benefits. Our results show that the additional encoding about anomaly occurrence boosts the autoencoder performance by bringing additional knowledge into the training process. Moreover, we point out that our encoder generalizes better because of the different training goals in a similar way penalty regularization does, improving the anomaly label and the encoding. Besides, we train our model on the anomaly free as well as on the anomalous training dataset. In addition, we enable transfer learning by sharing all hidden layers of the neural network and the autoencoder's encoder. 
	
	In many cases, neural network ensembles perform best by the combination of their individual predictions. The ensembles stamp out their weaknesses and bundle up their strengths. We create such ensemble benefits for our \textit{VAE SL} by implementing two methods for combining the predictions based on the reconstruction deviation between $x$ and $\tilde{x}$ as well as the direct anomaly label prediciton $\tilde{y}$, represented by the green arrows in Fig. \ref{figure:model}. The first method \textit{VAE SL Max} chooses the maximum out of both anomaly predictions. For the second method \textit{VAE SL Avg}, we include the distances of our anomaly predictions with respect to the decision border. For example, if a value of the prediction based on $x$ and $\tilde{x}$ is high above the decision point, we would credit higher certainty about this anomaly labeling compared to a value only a little above the decision point. For the anomaly label, instead of choosing the sampled $y$ that only contains zeros or ones, we take the mean $\pi$, which represents certainty, too. To combine both measurements and create an equal decision point of 0.5, we rescale the reconstruction deviation between 0 and 0.5 for data without anomaly and between 0.5 and 1 for anomalous data. Now, we can average the reconstruction deviation and the label prediction mean to our final anomaly measurement. If the average is above 0.5, we create labels $L$ for anomaly, otherwise we label the data as anomaly free.
	
	During the experiments, we detect some challenges in the \textit{VAE SL} models to overcome for optimization. We adjust the weighting of false positives compared to the unlabeled flawless data to counteract the imbalance of the labeled anomalies compared to false negatives for the loss function. Besides, we decrease the influence of the reconstruction probability because it repressed the other learning targets. To ensure a good bottleneck representation at the end of the training, we adjust the KL divergence to be comparatively small. This leads to the variances approaching zero at the beginning of the training. Certainly, it recovers when the other loss functions are getting small due to optimization. In the end, we have a satisfying bottleneck representation as well as good label prediction and data reconstruction.
	
	
	For the \textit{VAE MD} model, we extend the idea from our sparse anomaly label VAE to predict other sparse production metadata. Therefore, we repeat our extension procedure from \textit{VAE SL} for the additional knowledge. Hence, we change our architecture by adding another layer parallel to the bottleneck, concatenating it with the latent space for our decoder, and defining it as an output. We choose the activation and the anomaly prediction loss function based on the metadata type. E.g., continuous information is modeled as normal distribution, binary and categorical information with a sigmoid or softmax function and Bernoulli or Multinoulli distribution. We create a hyperparameter loss weight as well. This influences our model training in the same way the sparse anomaly labels do. Furthermore, we benefit from the new predictions directly by using them to improve our process and indirectly for our anomaly detection procedure. Our prediction's validity supports the anomaly labeling process by creating an additional anomaly measure to average it with the previous statements.
	
	We explain the details using the example from the "Sol" dataset. We include the sparse knowledge about the soldering machine's production status, which is grouped in 'production', 'equip', and 'rest'. First, we add a softmax layer of width three next to the bottleneck to concatenate it with the latent space sample. Next, we add a weighted Bernoulli loss function to our overall loss. After training, we use the new domain knowledge to evaluate the workload of the machine.
	 
	Moreover, we create a new anomaly measure. If our model cannot predict a single category with high certainty, we would suspect an anomalous point. For example, if the model predicts 'production' and 'rest' with an intermediate certainty, our process may be faulty. To transfer all categorical statements into a scale between zero and one, we think of all possible outcomes as barycentric coordinates with the edges $(1;0;0)$, $(0;1;0)$, $(0;0;1)$ as well as the centroid $(\frac{1}{3};\frac{1}{3};\frac{1}{3})$. We divide all values by the distance between the columns and the centroid to get an interval between 0 and 1. This anomaly measure is combined with the previous anomaly measures.
	
	We train our \textit{DNN} models on the supervised training data set. The overall structure constraints are derived from the autoencoder's encoder. We add a last layer with a sigmoid function and binary crossentropy loss for classification. 
	
	The anomaly detection based on a Principal Component Analysis reduces the input data into lower dimensional space with the number of principal components set as hyperparameter. After transforming our lower dimensional representation back to the original space, we classify every data point based on the reconstruction loss. We select the hyperparameters that results in the best f1- and AUC-score.
	
	For the Isolation Forest implementation, we randomly choose the number of base estimators from 50 to 400 and the contamination between 0 and 0.8 for 100 tries while selecting a fixed pseudo-random seed. We select the best search space combinations based on the highest accuracy, precision, and recall. Thereafter, we exceed our experiments in this direction several times, reducing the value range of the hyperparameters.
		
	\section{RESULTS}
	\label{RESULTS}
	
	After defining our experimental setup and models, we run experiments to evaluate the following questions:  
	
	\begin{enumerate}
		\item Are deep learning approaches worth the time and resources in comparison to other approaches?
		\item Are unsupervised or supervised learning techniques more effective for anomaly detection?
		\item Does probabilistic reconstruction significantly increase our model performance for VAEs?
		\item Can we benefit from combining supervised and unsupervised architectures?
		\item Does additional production metadata boost model performances?
		
	\end{enumerate}

	We test our seven model types on every data set to find the best combinations of accuracy, precision, and recall. Table \ref{Table:F1} and Table \ref{Table:AUC} contain the best f1- and AUC-scores out of 50 tries with the final model setups. We answer all our questions based on these summaries. A Friedman test significance with a $p$-value of 0.000159 (f1-scores) supports our results statistically, and \textit{VAE SL Avg} is statistically superior to many remaining methods under a two-sample t-tests. These $p$-values are also given in the tables.
		
	\begin{table}[h]
		\caption{F1-score model results and p-values of T-test applied to test datasets for anomaly detection}
		\label{Table:F1}
		\begin{center}
			\begin{tabular*}{\linewidth\setlength{\tabcolsep}{3.9pt}\setlength\arrayrulewidth{0.6pt}}{@{\extracolsep{\fill}}|c|c|c|c|c|c|c|c|c|}	
				\hline
				Model&Pneu&UPS&DB&Stor&Gen&Con&Sol&$p$-value\\
				\hline
				PCA&\cellcolor[rgb]{1,.64,0}0.5&\cellcolor[rgb]{.33,.54,0}0.94&\cellcolor[rgb]{.83,.62 0}0.74&\cellcolor[rgb]{1,.64,0}0.36&\cellcolor[rgb]{.51,.57,0}0.92&\cellcolor[rgb]{.51,.57,0}0.51&\cellcolor[rgb]{1,.64,0}0.15&0.05\\
				\hline
				IsoF&\cellcolor[rgb]{1,.64,0}0.54&\cellcolor[rgb]{.67,.6,0}0.87&\cellcolor[rgb]{.67,.6,0}0.84&\cellcolor[rgb]{.51,.57,0}0.39&\cellcolor[rgb]{1,.64,0}0.11&\cellcolor[rgb]{1,.64,0}0.23&\cellcolor[rgb]{1,.64,0}0.15&0.01\\
				\hline
				DNN&\cellcolor[rgb]{0,.5,0}\textbf{0.96}&\cellcolor[rgb]{.83,.62 0}0.83&\cellcolor[rgb]{0,.5,0}\textbf{0.94}&\cellcolor[rgb]{.33,.54,0}0.5&\cellcolor[rgb]{.51,.57,0}0.91& \cellcolor[rgb]{0,.5,0}0.76&\cellcolor[rgb]{0,.5,0}0.67& 0.52\\
				\hline
				VAE Err&\cellcolor[rgb]{.51,.57,0}0.95&\cellcolor[rgb]{.51,.57,0}0.89& \cellcolor[rgb]{1,.64,0}0.72&\cellcolor[rgb]{.67,.6,0}0.38&\cellcolor[rgb]{.51,.57,0}0.9&\cellcolor[rgb]{.51,.57,0}0.48&\cellcolor[rgb]{.83,.62 0}0.18 & 0.12\\
				\hline
				VAE Prob&\cellcolor[rgb]{.51,.57,0}0.94&\cellcolor[rgb]{0,.5,0}\textbf{0.96}&\cellcolor[rgb]{.51,.57,0}0.88&\cellcolor[rgb]{.83,.62 0}0.37&\cellcolor[rgb]{.51,.57,0}0.92&\cellcolor[rgb]{.83,.62 0}0.36&\cellcolor[rgb]{.51,.57,0}0.42&0.22\\
				\hline
				VAE SL Avg&\cellcolor[rgb]{0,.5,0}\textbf{0.96}&\cellcolor[rgb]{.33,.54,0}0.94&\cellcolor[rgb]{0,.5,0}\textbf{0.94}&\cellcolor[rgb]{0,.5,0}\textbf{0.62}&\cellcolor[rgb]{0,.5,0}\textbf{0.96}&\cellcolor[rgb]{0,.5,0}\textbf{0.83}&\cellcolor[rgb]{0,.5,0}\textbf{0.7}&opt\\
				\hline
				VAE SL Max&\cellcolor[rgb]{0,.5,0}\textbf{0.96}&\cellcolor[rgb]{1,.64,0}0.78&\cellcolor[rgb]{.33,.54,0}0.9& \cellcolor[rgb]{0,.5,0}0.59&\cellcolor[rgb]{.33,.54,0}0.93&\cellcolor[rgb]{0,.5,0}0.8&\cellcolor[rgb]{0,.5,0}\textbf{0.7}&0.57\\
				\hline
				VAE MD& n.a.& n.a.& n.a.& n.a.& n.a.& n.a.& \cellcolor[rgb]{.51,.57,0}0.4& \\
				\hline
			\end{tabular*}
		\end{center}
	\end{table}

	\begin{table}[h]
	\caption{AUC-score model results and p-values of T-test applied to test datasets for anomaly detection}
	\label{Table:AUC}
	\begin{center}
		\begin{tabular*}{\linewidth\setlength{\tabcolsep}{3.9pt}\setlength\arrayrulewidth{0.6pt}}{@{\extracolsep{\fill}}|c|c|c|c|c|c|c|c|c|}	
			\hline
			Model&Pneu&UPS&DB&Stor&Gen&Con&Sol&$p$-value\\
			\hline
			PCA& \cellcolor[rgb]{1,.64,0}0.56& \cellcolor[rgb]{.51,.57,0}0.91& \cellcolor[rgb]{1,.64,0}0.48& \cellcolor[rgb]{1,.64,0}0.51& \cellcolor[rgb]{.33,.54,0}0.98& \cellcolor[rgb]{.51,.57,0}0.72& \cellcolor[rgb]{1,.64,0}0.55&0.02\\
			\hline
			IsoF& \cellcolor[rgb]{.67,.6,0}0.62& \cellcolor[rgb]{1,.64,0}0.8& \cellcolor[rgb]{.67,.6,0}0.68& \cellcolor[rgb]{.51,.57,0}0.54& \cellcolor[rgb]{1,.64,0}0.83& \cellcolor[rgb]{1,.64,0}0.52& \cellcolor[rgb]{1,.64,0}0.54&0.001\\
			\hline
			DNN& \cellcolor[rgb]{.51,.57,0}0.97& \cellcolor[rgb]{1,.64,0}0.8& \cellcolor[rgb]{.33,.54,0}0.91& \cellcolor[rgb]{.33,.54,0}0.67& \cellcolor[rgb]{.67,.6,0}0.93& \cellcolor[rgb]{.33,.54,0}0.84& \cellcolor[rgb]{0,.5,0}\textbf{0.78}&0.29\\
			\hline
			VAE Err& \cellcolor[rgb]{.51,.57,0}0.97& \cellcolor[rgb]{.67,.6,0}0.84& \cellcolor[rgb]{.83,.62 0}0.53& \cellcolor[rgb]{.67,.6,0}0.53& \cellcolor[rgb]{0,.5,0}\textbf{0.99}& \cellcolor[rgb]{.51,.57,0}0.76& \cellcolor[rgb]{.83,.62 0}0.56&0.08\\
			\hline
			VAE Prob& \cellcolor[rgb]{.51,.57,0}0.97& \cellcolor[rgb]{.33,.54,0}0.95& \cellcolor[rgb]{.51,.57,0}0.85& \cellcolor[rgb]{.51,.57,0}0.54& \cellcolor[rgb]{0,.5,0}\textbf{0.99}& \cellcolor[rgb]{.67,.6,0}0.67& \cellcolor[rgb]{.51,.57,0}0.71&0.25\\
			\hline
			VAE SL Avg& \cellcolor[rgb]{0,.5,0}\textbf{0.98}& \cellcolor[rgb]{0,.5,0}\textbf{0.99}& \cellcolor[rgb]{0,.5,0}\textbf{0.95}& \cellcolor[rgb]{0,.5,0}\textbf{0.73}& \cellcolor[rgb]{.51,.57,0}0.97& \cellcolor[rgb]{0,.5,0}\textbf{0.94}& \cellcolor[rgb]{.33,.54,0}0.77&opt\\
			\hline
			VAE SL Max& \cellcolor[rgb]{0,.5,0}\textbf{0.98}& \cellcolor[rgb]{.33,.54,0}0.94& \cellcolor[rgb]{.33,.54,0}0.93& \cellcolor[rgb]{.33,.54,0}0.72& \cellcolor[rgb]{.67,.6,0}0.95& \cellcolor[rgb]{.33,.54,0}0.88& \cellcolor[rgb]{.33,.54,0}0.77&0.68\\
			\hline
			VAE MD& n.a.& n.a.& n.a.& n.a.& n.a.& n.a.& \cellcolor[rgb]{1,.64,0}0.63&\\
			\hline
		\end{tabular*}
	\end{center}
	\end{table}

	We observe that neural network-based anomaly detection methods outperform the classic approaches Principal Component Analysis and Isolation Forest on every dataset. Deep learning enables training more complex structures that can model the production processes more detailed. Moreover, recurrent parts in neural networks enable to model timing-based process behavior. Nevertheless, classic models \textit{PCA} and \textit{IsoF} score more than 90 percent for 2 of our 7 scenarios. We suggest using standard approaches as starting point to elaborate the dataset's properties and specify the hyperparameter ranges for neural network training.
	 
	In our experiments, Classifying Neural Networks \textit{DNN} improve on vanilla Variational Autoencoders \textit{VAE Err} and \textit{VAE Prob} in 5 out of 7 times. This emphasizes the importance of labeling data for anomaly detection methods. For our experiments, we generate the supervised training and testing dataset out of the same data records. That influences our supervised model performance because every type of anomaly was present in the training data. Besides, a Classifying Neural Network gives us insights into how our anomalies are encoded in our process data.
	
	If we aim to improve our \textit{VAE Err}, using reconstruction probability is highly beneficial for anomaly detection. \textit{VAE Prob} models with probabilistic reconstruction get better results on 4 and similar results on 2 out of 7 datasets. They also create higher reconstruction losses for almost every anomaly. The only exception is the "Con" dataset. We are not able to create a satiable reconstruction probability VAE. We suspect that finding a suitable model becomes too complicated in our experimental setup. However, we should find an equal or better performing model because the reconstruction error is a special case of the reconstruction probability.
	
	We investigate the performance of our sparse label extension \textit{VAE SL}. Taking f1-scores into account, the averaging prediction \textit{VAE SL Avg} model is better in 4 and equal in 2 of 7 experiments in comparison to all other models. Comparing AUC-scores and introducing sparse label information obtains better results in 5 and similar results in 1 of 7 compared to our other models. In conclusion, the combination of supervised and unsupervised methods leads to a neural network that outperforms both groups. We credit the following points for the excellent performance: First, the model can train on the supervised and unsupervised data. More data often leads to better generalization. Second, neural network ensembles tend to be the right choice for the best overall result. Our model can be interpreted as a model ensemble with parameter sharing. Third, we assume transfer learning between creating a latent space and the anomaly label in the bottleneck. Moreover, the reconstruction benefits from the predicted anomaly label. 
	
	We recognize less performance in 5 or equal performance in 2 cases for our prediction \textit{VAE SL Max} compared to the averaging combination method \textit{VAE SL Avg}. In comparison to the other methods, the model performs better in 3, equal in 2, and worse in 2 experiments. This reveals the importance of the correct prediction combination method. Even if we have a favorite for our experiments, the correct combination procedure depends on the training goal definition. If we add weight to our detection's specificity, \textit{VAE SL Max} combinations can be the right choice.
	
	Because the inclusion of additional process knowledge boosts our model performances, we evaluate our results for the production metadata model \textit{VAE MD} on the "Sol" dataset. Unfortunately, our model cannot beat our \textit{VAE SL Avg} models, having a difference of 0.3 in f1- and 0.14 in AUC-score. Because the model should at least perform equally compared to the sparse label VAE, we state the following hypothesis: The occurrence of an anomaly is not encoded in the deviation between the predicted and the real production status. 
		
	\section{CONCLUSION}
	\label{CONCLUSION}
	
	We aim to find a strategy that includes machine learning to enable data-driven anomaly detection supported by the fourth industrial revolution's data transparency. In the end, the efforts for feature and model engineering pay back because of the automated anomaly detection process that decreases downtimes. Additionally, anomaly detection is a method to enable predictive maintenance and autonomous optimization for smart factories. Moreover, deep learning is worth investigating feature and model engineering because of higher performances for anomaly labeling.
	
	Talking about training targets, supervised and unsupervised neural networks have benefits and drawbacks that show up on our datasets. Supervised learning can fail on unseen types of anomalies, while unsupervised deviation-based methods suffer from compromising between generalization and specialization. We always investigate both methods in parallel to find out how well the anomalies are encoded into the process data. Moreover, we create a model that combines the benefits and eliminates the drawbacks of both methods. 
	
	In summary, we find strategies to include sparse production knowledge to increase anomaly detection reliability. Using reconstruction probability on the decoder's outputs boosts all our VAE models to perform better than reconstruction error. Our findings suggest better performance by bringing probabilistic approaches into the network training and detection process. Furthermore, the sparse anomaly label extension of the Variational Autoencoders leads to better network performances based on more training data availability, transfer learning, and the ensemble effect.
	
	We can include other production metadata in the same way we include the supervised anomaly knowledge. The VAE encoder's extension leads to useful predictions of the "Sol" dataset's production status. Unfortunately, we did not find a way to derive anomalies out of the status predictions to boost our overall detection mechanism. Nevertheless, we can use the additional knowledge about the production for further investigations. In summary, our models can handle flexible information flows based on the ability to include sparse labels for model training.	
	\\
	
	\addtolength{\textheight}{-12cm}   
	
		
	\bibliographystyle{ieeetran}
	\bibliography{ieeetran}

\begin{thebibliography}{10}
\providecommand{\url}[1]{#1}
\csname url@samestyle\endcsname
\providecommand{\newblock}{\relax}
\providecommand{\bibinfo}[2]{#2}
\providecommand{\BIBentrySTDinterwordspacing}{\spaceskip=0pt\relax}
\providecommand{\BIBentryALTinterwordstretchfactor}{4}
\providecommand{\BIBentryALTinterwordspacing}{\spaceskip=\fontdimen2\font plus
\BIBentryALTinterwordstretchfactor\fontdimen3\font minus
  \fontdimen4\font\relax}
\providecommand{\BIBforeignlanguage}[2]{{%
\expandafter\ifx\csname l@#1\endcsname\relax
\typeout{** WARNING: IEEEtran.bst: No hyphenation pattern has been}%
\typeout{** loaded for the language `#1'. Using the pattern for}%
\typeout{** the default language instead.}%
\else
\language=\csname l@#1\endcsname
\fi
#2}}
\providecommand{\BIBdecl}{\relax}
\BIBdecl

\bibitem{Daily.2017}
M.~Daily, S.~Medasani, R.~Behringer, and M.~Trivedi, ``Self-{D}riving {C}ars,''
  \emph{Computer}, vol.~50, no.~12, pp. 18--23, 2017.

\bibitem{Antonopoulos.2010}
\BIBentryALTinterwordspacing
N.~Antonopoulos and L.~Gillam, Eds., \emph{Cloud {C}omputing}.\hskip 1em plus
  0.5em minus 0.4em\relax London: {Springer London}, 2010. [Online]. Available:
  \url{http://dx.doi.org/10.1007/978-1-84996-241-4}
\BIBentrySTDinterwordspacing

\bibitem{HenningKagermannJohannesHelbigArianeHellingerWolfgangWahlster.2013}
\BIBentryALTinterwordspacing
{Henning Kagermann, Johannes Helbig, Ariane Hellinger, Wolfgang Wahlster},
  \emph{Recommendations for implementing the strategic initiative INDUSTRIE
  4.0}, 2013. [Online]. Available:
  \url{https://en.acatech.de/publication/recommendations-for-implementing-the-strategic-initiative-industrie-4-0-final-report-of-the-industrie-4-0-working-group}
\BIBentrySTDinterwordspacing

\bibitem{A.Alvanpour.2020}
{A. Alvanpour}, {S. K. Das}, {C. K. Robinson}, {O. Nasraoui}, and {D. Popa},
  ``Robot {F}ailure {M}ode {P}rediction with {E}xplainable {M}achine
  {L}earning,'' in \emph{2020 IEEE CASE Conference}, 2020, pp. 61--66.

\bibitem{Omar.2013}
S.~Omar, A.~Ngadi, and H.~H. Jebur, ``Machine {L}earning {T}echniques for
  {A}nomaly {D}etection: {A}n {O}verview,'' \emph{IJCA}, vol.~79, no.~2, 2013.

\bibitem{Z.Wang.2020}
{Z. Wang}, {G. Xu}, {J. Wang}, {M. Liu}, and {Y. Ma}, ``Cross-{D}omain {F}ault
  {D}iagnosis with {O}ne-{D}imensional {C}onvolutional {N}eural {N}etwork,'' in
  \emph{2020 IEEE CASE Conference}, 2020, pp. 494--499.

\bibitem{VansonBourne.2020}
\BIBentryALTinterwordspacing
{Vanson Bourne}, ``After the fall: The {C}osts, {C}auses {\&} {C}onsequences of
  {U}nplanned {D}owntimes.'' [Online]. Available:
  \url{https://lp.servicemax.com/Vanson-Bourne-Whitepaper-Unplanned-Downtime-LP.html}
\BIBentrySTDinterwordspacing

\bibitem{Wang.2018}
J.~Wang, Y.~Ma, L.~Zhang, R.~X. Gao, and D.~Wu, ``Deep {L}earning for {S}mart
  {M}anufacturing: {M}ethods and {A}pplications,'' \emph{Journal of
  Manufacturing Systems}, vol.~48, pp. 144--156, 2018.

\bibitem{Goernitz.2013}
N.~Goernitz, M.~Kloft, K.~Rieck, and U.~Brefeld, ``Toward {S}upervised
  {A}nomaly {D}etection,'' \emph{Journal of Artificial Intelligence Research},
  vol.~46, pp. 235--262, 2013.

\bibitem{Berkhahn.2019}
\BIBentryALTinterwordspacing
F.~Berkhahn, R.~Keys, W.~Ouertani, N.~Shetty, and D.~Gei{\ss}ler, ``Augmenting
  {V}ariational {A}utoencoders with {S}parse {L}abels: {A} {U}nified
  {F}ramework for {U}nsupervised, {S}emi-(un)supervised, and {S}upervised
  {L}earning.'' [Online]. Available: \url{https://arxiv.org/pdf/1908.03015}
\BIBentrySTDinterwordspacing

\bibitem{Hansen.1990}
\BIBentryALTinterwordspacing
L.~K. Hansen and P.~Salamon, ``Neural {N}etwork {E}nsembles,'' \emph{IEEE
  Transactions on Pattern Analysis and Machine Intelligence}, 1990. [Online].
  Available: \url{http://dx.doi.org/10.1109/34.58871}
\BIBentrySTDinterwordspacing

\bibitem{Doersch.2016}
\BIBentryALTinterwordspacing
C.~Doersch, ``Tutorial on variational autoencoders.'' [Online]. Available:
  \url{https://arxiv.org/pdf/1606.05908}
\BIBentrySTDinterwordspacing

\bibitem{Goh.2017}
J.~Goh, S.~Adepu, M.~Tan, and Z.~S. Lee, ``Anomaly {D}etection in {C}yber
  {P}hysical {S}ystems {U}sing {R}ecurrent {N}eural {N}etworks,'' in \emph{IEEE
  HASE Symposium}.\hskip 1em plus 0.5em minus 0.4em\relax Piscataway, NJ: IEEE,
  2017, pp. 140--145.

\bibitem{Liu.2008}
F.~T. Liu, K.~M. Ting, and Z.-H. Zhou, ``Isolation {F}orest,'' in \emph{Eighth
  IEEE international conference on data mining}, 2008, pp. 413--422.

\bibitem{Wold.1987}
S.~Wold, K.~Esbensen, and P.~Geladi, ``Principal {C}omponent {A}nalysis,''
  \emph{Chemometrics and intelligent laboratory systems}, vol.~2, no. 1-3, pp.
  37--52, 1987.

\bibitem{An.2015}
J.~An and S.~Cho, ``Variational {A}utoencoder based {A}nomaly {D}etection using
  {R}econstruction {P}robability,'' \emph{Special Lecture on IE}, vol.~2,
  no.~1, pp. 1--18, 2015.

\bibitem{Hawkins.1980}
D.~M. Hawkins, \emph{Identification of Outliers}, ser. Monographs on Applied
  Probability and Statistics.\hskip 1em plus 0.5em minus 0.4em\relax Dordrecht:
  Springer, 1980.

\bibitem{Muller.2020}
A.~M{\"u}ller, M.~Lange-Hegermann, and A.~v. Birgelen, ``Automatisches
  {T}raining eines {V}ariational {A}utoencoders f{\"u}r {A}nomalieerkennung in
  {Z}eitreihen,'' in \emph{Automation 2020}.\hskip 1em plus 0.5em minus
  0.4em\relax D{\"u}sseldorf: {VDI Verlag}, 2020, pp. 687--698.

\bibitem{Niggemann.2018}
O.~Niggemann and P.~Sch{\"u}ller, Eds., \emph{IMPROVE - Innovative Modelling
  Approaches for Production Systems to Raise Validatable Efficiency}.\hskip 1em
  plus 0.5em minus 0.4em\relax {Springer Berlin Heidelberg}, 2018, vol.~8.

\bibitem{Schuster.2018}
\BIBentryALTinterwordspacing
R.~Schuster, N.~Moriz, and A.~von Birgelen, ``Genesis {D}emonstrator {D}ata for
  {M}achine {L}earning.'' [Online]. Available:
  \url{https://www.kaggle.com/inIT-OWL/genesis-demonstrator-data-for-machine-learning}
\BIBentrySTDinterwordspacing

\bibitem{Iosbina.2020}
\BIBentryALTinterwordspacing
F.~Iosb-ina, ``Dataset: Industry,'' 2020. [Online]. Available:
  \url{https://www.kaggle.com/smartfactoryowl/industry}
\BIBentrySTDinterwordspacing

\end{thebibliography}

\end{document}